\documentclass[journal]{IEEEtran}
\usepackage{graphicx}
\usepackage{cite}
\usepackage{amssymb}
\usepackage{amsmath}
\usepackage{enumerate}
\usepackage{booktabs}
\usepackage{algorithm}
\usepackage{algorithmic}
\usepackage{color}
\usepackage{colortbl}
\usepackage[colorlinks,linkcolor=red,anchorcolor=blue,citecolor=green]{hyperref}
\usepackage{bbm}

\usepackage{multirow}
\usepackage[table,xcdraw]{xcolor}
\begin{document}

\title{Txt2Img-MHN: Remote Sensing Image Generation from Text Using Modern Hopfield Networks}

\author{Yonghao~Xu,~\IEEEmembership{Member,~IEEE,} Weikang~Yu,~\IEEEmembership{Student Member,~IEEE,}~Pedram~Ghamisi,~\IEEEmembership{Senior Member,~IEEE},\\~Michael~Kopp,~\IEEEmembership{Member,~IEEE,}~and~Sepp~Hochreiter
\thanks{Y. Xu, P. Ghamisi, M. Kopp, and S. Hochreiter are with the Institute of Advanced Research in Artificial Intelligence (IARAI), 1030 Vienna, Austria (e-mail: yonghao.xu@iarai.ac.at; pedram.ghamisi@iarai.ac.at; michael.kopp@iarai.ac.at; sepp.hochreiter@iarai.ac.at).}
\thanks{Y. Xu is also with Computer Vision Laboratory, Department of Electrical Engineering, Link\"{o}ping University, 58183 Link\"{o}ping, Sweden (e-mail: yonghao.xu@liu.se).}
\thanks{W. Yu is with Helmholtz-Zentrum Dresden-Rossendorf, Helmholtz Institute Freiberg for Resource Technology, Machine Learning Group, 09599 Freiberg, Germany (e-mail: w.yu@hzde.de).}
\thanks{P. Ghamisi is also with Helmholtz-Zentrum Dresden-Rossendorf, Helmholtz Institute Freiberg for Resource Technology, Machine Learning Group, 09599 Freiberg, Germany (e-mail: p.ghamisi@hzdr.de).}
\thanks{S. Hochreiter is also with ELLIS Unit Linz and LIT AI Lab, Institute for Machine Learning, Johannes Kepler University, 4040 Linz, Austria (e-mail: hochreit@ml.jku.at).}
}

\markboth{IEEE Transactions on Image Processing, October~2023}%
{Shell \MakeLowercase{\textit{et al.}}: Bare Demo of IEEEtran.cls for IEEE Journals}

\maketitle

\begin{abstract}
The synthesis of high-resolution remote sensing images based on text descriptions has great potential in many practical application scenarios. Although deep neural networks have achieved great success in many important remote sensing tasks, generating realistic remote sensing images from text descriptions is still very difficult. To address this challenge, we propose a novel text-to-image modern Hopfield network (Txt2Img-MHN). The main idea of Txt2Img-MHN is to conduct hierarchical prototype learning on both text and image embeddings with modern Hopfield layers. Instead of directly learning concrete but highly diverse text-image joint feature representations for different semantics, Txt2Img-MHN aims to learn the most representative prototypes from text-image embeddings, achieving a coarse-to-fine learning strategy. These learned prototypes can then be utilized to represent more complex semantics in the text-to-image generation task. To better evaluate the realism and semantic consistency of the generated images, we further conduct zero-shot classification on real remote sensing data using the classification model trained on synthesized images. Despite its simplicity, we find that the overall accuracy in the zero-shot classification may serve as a good metric to evaluate the ability to generate an image from text. Extensive experiments on the benchmark remote sensing text-image dataset demonstrate that the proposed Txt2Img-MHN can generate more realistic remote sensing images than existing methods. Code and pre-trained models are available online (https://github.com/YonghaoXu/Txt2Img-MHN).

\end{abstract}


\IEEEpeerreviewmaketitle

\section{Introduction}

\IEEEPARstart{W}{ith} the rapid development of imaging and aerospace technology, a massive amount of remote sensing data collected by airborne or spaceborne sensors is now available \cite{ghamisi2017advanced}. While the explosive growth of remote sensing data provides a great opportunity for researchers to explore and monitor the Earth, it also brings about a challenge: to accurately interpret and understand the abundant semantic information in these data \cite{uaers,zhang2022artificial}. Currently, most of the existing research for the interpretation of remote sensing data focuses on tasks like scene classification \cite{cheng2020remote,adv_rs,potnis2021semantics}, object detection \cite{sun2021pbnet,ghorbanzadeh2022landslide4sense}, and semantic segmentation \cite{ding2020lanet,xu2022consistency}. Although many machine learning models, especially the deep learning-based ones, have been developed to obtain high accuracy in these tasks, these models have shown limited success in generalizing beyond their limited task and are often adversely affected by domain shifts due to different imaging environments \cite{zhang2016deep,maggiori2017can}. One of the main reasons lies in the fact that these models are designed to be task-specific and they can only interpret the remote sensing image from a specific view, while ignoring the high-level semantic relationship between different ground objects \cite{lobry2020rsvqa}. Take the object detection task, for example. State-of-the-art object detection models usually can accurately detect the locations of specific objects of interest like airplanes or vehicles, but can not understand high-level semantic concepts like the number of objects or the spatial distribution attributes of different objects. As a result, a more accurate and comprehensive interpretation of remote sensing data still depends largely on expert knowledge \cite{yuan2022remote}.

\begin{figure}
  \centering
  \includegraphics[width=\linewidth]{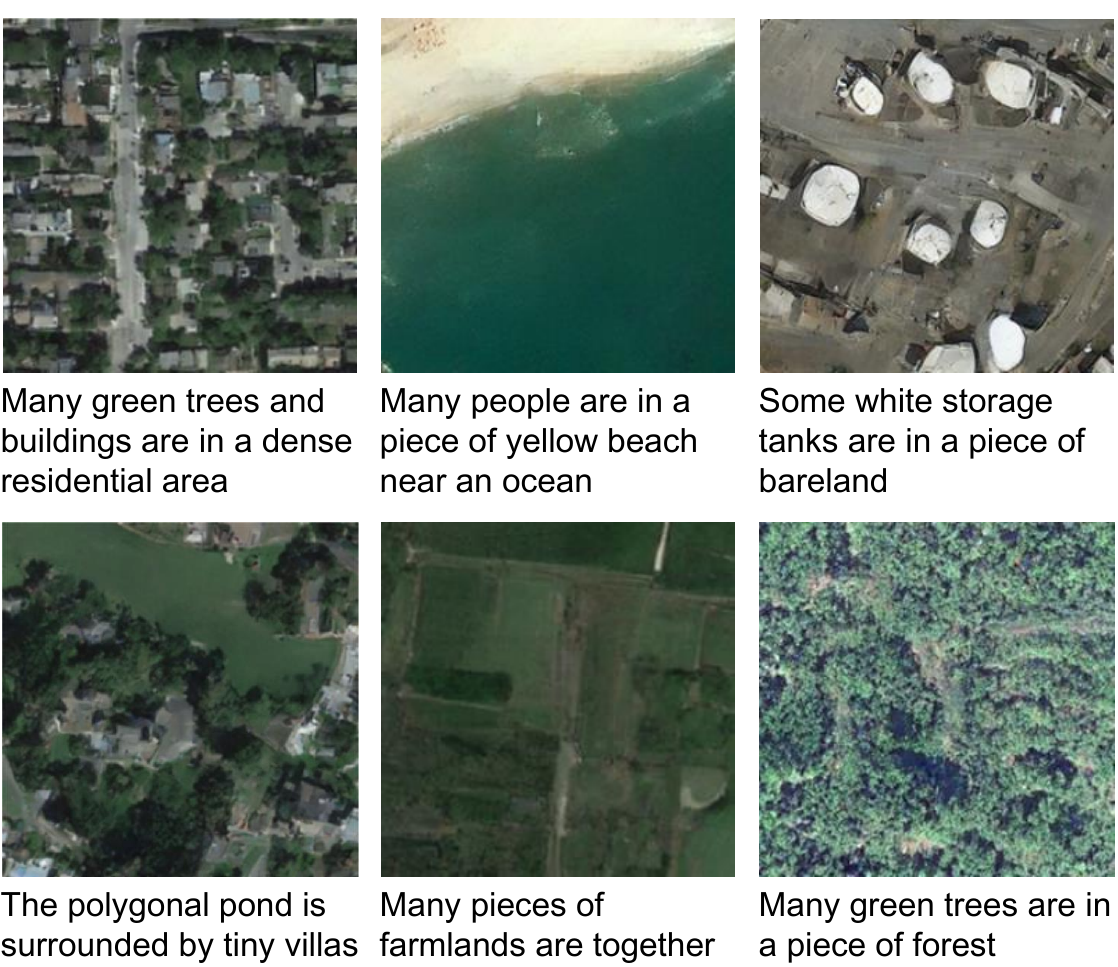}
  \caption{Examples of the remote sensing images generated from the given text descriptions using the proposed text-to-image modern Hopfield network (Txt2Img-MHN).}
\label{fig:intro}
\end{figure}

To gain deeper insights into the semantic information in remote sensing images, recent research attempts have been made to conduct image captioning \cite{zhang2019description,li2020multi,liu2022remote} and image change captioning \cite{hoxha2022change,liu2022remote_change,chang2023changes} for remote sensing data. These approaches aim to summarize the semantic content or the land cover changes of the given images through descriptive text sentences. In \cite{shi2017can}, Shi et al. propose a multi-stage image captioning framework using fully convolutional networks (FCNs), which can extract multi-level semantics from the input remote sensing image. Once the key ground objects and scene-level semantic information are detected, the framework can then generate language descriptions by integrating the hierarchical features obtained from the previous stage. Lu et al. further collect a benchmark remote sensing text-image dataset named RSICD (Remote Sensing Image Captioning Dataset) \cite{lu2017exploring}. They also conduct a comprehensive review of encoder-decoder-based image captioning methods.

So far, researchers have put much effort into the remote sensing image captioning task \cite{li2020truncation,hoxha2021novel,zia2022transforming}, while the inverse problem, text-to-image generation, has rarely been studied in the remote sensing community. In contrast to image captioning, text-to-image generation aims to generate realistic images based on the given text descriptions. Despite its difficulty, there is great potential to apply this technique in real application scenarios:
\begin{itemize}
    \item \textit{Simulated urban planning:} With the text-to-image generation technique, researchers can synthesize realistic scenes for the region of interest using text descriptions from urban planners. The simulated remote sensing images may help urban planners better evaluate whether the current designs are feasible, assisting the final decisions.
    \item \textit{Data augmentation:} Although there is a massive amount of remote sensing data available, acquiring high-quality labels for them is still very challenging and time-consuming. With the text-to-image generation technique, researchers can derive realistic remote sensing images from text descriptions. Since the text descriptions usually contain the keywords of the main object or scene of the generated image, it would be effortless to obtain its corresponding annotation, and thereby mitigate the insufficiency of labeled samples \cite{chen2021remote,singh2021sigan,kim2022gan}.
\end{itemize}

In \cite{bejiga2019retro}, Bejiga et al. introduced the text-to-image generation task into the remote sensing community for the first time, adopting a generative adversarial network (GAN) to synthesize gray-scale remote sensing images using ancient text descriptions. Chen et al. further proposed the text-based deeply-supervised GAN to synthesize satellite images with a spatial size of $128\times 128$ \cite{chen2021remote}. The generated images are then utilized to augment the training set for the change detection task. Considering that the structure information is an important factor in evaluating the fidelity of the generated images, Zhao et al. propose the structured GAN to synthesize remote sensing images from text with a spatial size of $256\times 256$ \cite{zhao2021text}. However, due to the complex spatial distribution of different ground objects and the huge modality gap between text descriptions and remote sensing images, the performance of the aforementioned methods is limited.

To generate more realistic remote sensing images, we propose the text-to-image modern Hopfield network (Txt2Img-MHN). The key idea of Txt2Img-MHN is to conduct hierarchical prototype learning on both text and image embeddings using a Hopfield Lookup via a `HopfieldLayer' as described in \cite{Ramsauer:21}. Instead of directly learning concrete but highly diverse text-image joint feature representations for different semantics, which is more difficult, Txt2Img-MHN aims to learn the most representative prototypes from all text-image embeddings. The learned prototypes can then be further utilized to represent more complex semantics in the text-to-image generation task. Fig. \ref{fig:intro} provides some examples of the remote sensing images generated from the given text descriptions using the proposed Txt2Img-MHN.

The main contributions of this paper are summarized as follows.

\begin{enumerate}
\item We propose the text-to-image modern Hopfield network (Txt2Img-MHN) based on learned Hopfield Lookups in layers of type `Hopfieldlayer' and the self-attention module for remote sensing image generation from text. Extensive experiments on the benchmark remote sensing text-image dataset demonstrate that the proposed Txt2Img-MHN can generate more realistic remote sensing images than the existing methods.
\item We comprehensively compare and analyze the performance of the vector quantized variational autoencoder (VQVAE) and vector quantized generative adversarial network (VQGAN) for text-to-image generation in the remote sensing scenario, which may bring about new insight for researchers to better design image encoders/decoders for remote sensing data.
\item We further conduct zero-shot classification on real remote sensing data using the classification model trained on synthesized images. Compared to traditional evaluation metrics like Inception Score, we find the overall accuracy in the zero-shot classification may serve as a better criterion for the text-to-image generation task, especially from the perspective of semantic consistency.
\end{enumerate}

The rest of this paper is organized as follows. Section II reviews works related to this study. Section III describes the proposed Txt2Img-MHN in detail. Section IV presents the information on datasets used in this study and the experimental results. Conclusions and other discussions are summarized in Section V.

\section{Related Work}
This section makes a brief review of the existing text-to-image generation and prototype learning methods, along with the modern Hopfield networks.
\subsection{Text-to-Image Generation}
Text-to-image generation is an important and challenging task, which aims to generate realistic images according to the given natural language descriptions. The early research on text-to-image generation mainly focuses on GAN-based methods \cite{reed2016generative,reed2016learning,zhang2017stackgan}. The first pioneering work is text-conditional GAN, where the generator is designed to synthesize realistic images based on the extracted text features and cheat the discriminator, while the discriminator, on the other hand, aims to distinguish whether the input image is real or fake \cite{reed2016generative}. However, this method can only generate images with a spatial size of $64\times 64$. To generate images with a higher spatial resolution, researchers have also considered adding the key point locations as the auxiliary input for the generator \cite{reed2016learning}. Based on this idea, Reed et al. propose the generative adversarial what-where network, which can synthesize images with a spatial size of $128\times 128$. Another interesting work is StackGAN, where the authors propose to synthesize images with the stacked generator \cite{zhang2017stackgan}. While the output of the first generator is low quality, with a spatial size of only $64\times 64$, the second generator will receive it as new input and refine the synthesized images to $256\times 256$.

Apart from the GAN-based methods, recent research focuses on the transformer-based models. One of the most representative work is DALL-E, which is a variant of GPT-3 (generative pre-trained transformer-3) \cite{brown2020language} with 12 billion parameters \cite{ramesh2021zero}. Owing to the powerful learning ability of the giant transformer model and the large-scale training data (250 million text-image pairs), DALL-E is able to combine unrelated language concepts plausibly and synthesize high-quality images with a spatial size of $256\times 256$. Very recently, the second generation of DALL-E has been published \cite{ramesh2022hierarchical}. Compared to DALL-E 1, DALL-E 2 is a more advanced artificial intelligence system that directly learns the relationship between images and text descriptions like the CLIP (contrastive language–image pre-training) model \cite{radford2021learning} does. It can generate realistic images and even artistic work with a spatial resolution of $1024\times 1024$ based on the given text descriptions using the diffusion model \cite{ramesh2022hierarchical}.

So far, most of the research focuses on the generation of natural images, while the related works on the remote sensing community are still immature due to the complex spatial distribution of different ground objects and the huge modality gap between text descriptions and remote sensing images \cite{bejiga2019retro,chen2021remote,zhao2021text}. Since the application scenarios of remote sensing tasks generally need higher requirements for the fidelity and plausibility of the synthesized images, the challenge of generating realistic remote sensing images from the text descriptions still remains.

\subsection{Prototype Learning}
Prototype learning is a commonly used technique in computer vision and pattern recognition tasks that aims to optimize or select the most representative anchors or data points from the training samples \cite{liu2001evaluation}. One of the most well-known prototype learning algorithms is the k-nearest neighbor (KNN) \cite{dudani1976distance}, where the prototypes are determined by choosing the nearest neighbors from all training samples with Euclidean distance. Since directly selecting prototypes from the complete training set is very time-consuming and entails heavy storage burdens, the LVQ (learning vector quantization) algorithm is proposed \cite{kohonen1990self}. Specifically, LVQ improves prototype learning in the 1-NN classifier by taking the class boundaries of those hard categories into consideration; the input testing samples can be classified with only a small number of prototypes without visiting all training samples.

With the advent of the deep learning era, recent research focuses on learning the prototypes automatically through the use of well-designed neural networks \cite{li2021adaptive}. In \cite{yang2018robust}, Yang et al. propose the convolutional prototype learning framework for the image recognition task, where they simultaneously optimize the convolutional neural network (CNN) and the prototypes in different categories. They find that using prototype matching for decision-making helps to improve the robustness of the classification. Dong et al. further introduce the prototype learning technique into the few-shot semantic segmentation task, using a sub-network to learn prototypes from the supporting set \cite{dong2018few}. Other successful applications include action recognition \cite{wang2021interactive}, and face recognition \cite{deng2021variational}.

Unlike the aforementioned methods, in this study, we propose to conduct the prototype learning from text-image embeddings using modern Hopfield networks \cite{Ramsauer:21}. While directly learning highly diverse text-image joint feature representations for different semantics is very difficult, the proposed method could simplify the task by first learning the most representative prototypes, which can be further utilized to represent more complex semantics.

\subsection{Modern Hopfield Networks}
Associative memory networks are a type of neural network architecture designed for pattern recognition and recall tasks \cite{mceliece1987capacity}. They are based on the idea of ``associative memory'', which is the ability to retrieve a stored memory based on partial or incomplete input. As the most representative energy-based, binary associative memory networks, Hopfield networks popularized artificial neural networks in the 1980s \cite{Hopfield:82,Hopfield:84}. Since they have a finite storage capacity, the classical binary Hopfield networks can only store a limited number of patterns. An interesting finding is that the storage capacity of classical binary Hopfield networks can be considerably increased by polynomial or exponential terms in the energy function \cite{Krotov:16,Demircigil:17}. In contrast to these binary memory networks, Ramsauer et al. proposed modern Hopfield networks which achieve continuous associative memory. These modern Hopfield networks for deep learning architectures have an energy function with continuous states, which can store exponentially many samples that can be retrieved with one update only \cite{Ramsauer:21}. Crucially, they are differentiable and can thus be embedded in deep learning architectures trained by gradient descent. Modern Hopfield networks have already been successfully applied to immune repertoire classification \cite{widrich2020modern} and chemical reaction prediction \cite{Seidl:22}.

\begin{figure*}
  \centering
  \includegraphics[width=\linewidth]{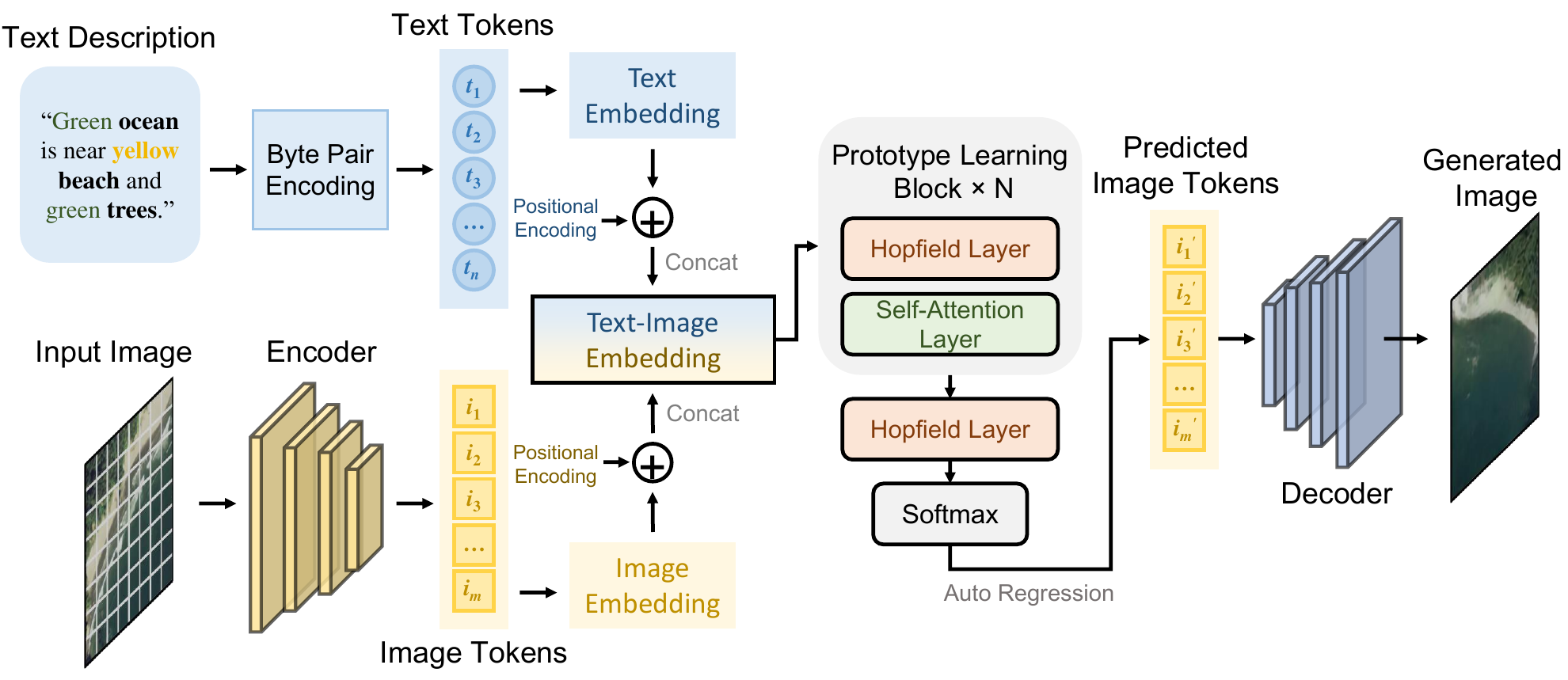}
  \caption{Illustration of the proposed text-to-image modern Hopfield network (Txt2Img-MHN). Given a text-image pair, we first use the BPE (byte pair encoding) encoder and the image encoder to obtain the corresponding text and image tokens, which are further transformed into the text-image embeddings. These embeddings are then fed into the proposed prototype learning blocks ($\times N$) consisting of learned Hopfield Lookups in a `HopfieldLayer' followed by a self-attention layer, followed by another such `HopfieldLayer' and the softmax function to obtain the predicted image tokens using autoregression. Finally, the predicted image tokens are fed into the image decoder to synthesize realistic remote sensing images. The whole network is optimized by minimizing the cross-entropy loss between the predicted image tokens and the original input image tokens.}
\label{fig:mhn}
\end{figure*}

\section{Methodology}
While most of the existing text-to-image generation research in the remote sensing field focuses on GAN-based methods \cite{bejiga2019retro,chen2021remote,zhao2021text}, we propose a prototype learning-based approach: Text-to-image modern Hopfield network (Txt2Img-MHN). The key ingredient for our method is the modern Hopfield network introduced in \cite{Ramsauer:21}, which is an associative memory of continuous patterns that can store exponentially many such patterns, can retrieve patterns in one update step, and is differentiable. The last property means that it can be inserted into any deep learning architectures trained by gradient descent  as in \cite{Seidl:22, furst2021cloob}. The key idea of Txt2Img-MHN is to conduct hierarchical prototype learning on the text-image embeddings through the use of learned queries via modern Hopfield network layers akin to \cite{widrich2020modern}. Compared to learning concrete but highly diverse semantics directly from the input text-image pairs, the prototype learning strategy used in this study can achieve the coarse-to-fine learning purpose, in which only the most representative prototypes are learned. These learned prototypes can then be utilized to represent more complex semantics in the text-to-image generation task, achieving state-of-the-art performance.

\subsection{Overview of the Proposed Txt2Img-MHN}
The flowchart of the proposed Txt2Img-MHN is shown in Fig. \ref{fig:mhn}. In the training phase, given a text-image pair, we first use the BPE (byte pair encoding) encoder \cite{sennrich2015neural} to separate the words in the text description and obtain the corresponding tokens. Formally, let $\boldsymbol{t}=\left(t_1,t_2,\cdots,t_n\right)$ denote the obtained text tokens, where $n$ is the maximum length of the input sentence. In cases where the number of words in the input text description is smaller than $n$, we use $0$ as the placeholder to fill the empty tokens. We tokenize the input image $x$ by adopting a pre-trained image encoder (see Section \ref{section:vqvae} for details) to tokenize it. Let $\boldsymbol{i}=\left(i_1,i_2,\cdots,i_m\right)$ denote the obtained image tokens, where $m$ is the total length of the image tokens. Similar to the Transformer model \cite{vaswani2017attention}, for both text and image tokens, we use an embedding layer with positional encoding to transform them into the embedding space with a dimension of $d_{emb}$. Here, positional encoding can help to inject the relative or absolute position information among different tokens in the sequence \cite{gehring2017convolutional}. Let ${\rm emb}_{text}\in \mathbb{R}^{n\times d_{emb}}$ and ${\rm emb}_{img}\in \mathbb{R}^{m\times d_{emb}}$ denote the obtained text embedding and image embedding, respectively. The joint text-image embedding features ${\rm emb}_{text-img}$ can be obtained by:
\begin{equation}
{\rm emb}_{text-img}=\left[{\rm emb}_{text};{\rm emb}_{img}\right]\in \mathbb{R}^{(n+m)\times d_{emb}}.
\label{eq:emb}
\end{equation}

${\rm emb}_{text-img}$ is then fed into the proposed prototype learning blocks ($\times N$) consisting of a HopfieldLayer each learning Hopfield Lookups followed by a self-attention layer, followed by another such Hopfieldlayer and the softmax function leading to the predicted image tokens $\boldsymbol{i'}=\left(i_1',i_2',\cdots,i_m'\right)$. The whole network is optimized by minimizing the cross-entropy loss between the probability vector of the predicted image tokens and the original image tokens.

In the test phase, since there are only text descriptions as input, the predicted image tokens will be obtained auto-regressively \cite{graves2013generating}. Specifically, given the text tokens $\boldsymbol{t}=\left(t_1,t_2,\cdots,t_n\right)$, the network will predict the first image token $i_1'$ at the first iteration (in the practical implementation, we add $t_0=0$ at the beginning of the text tokens to achieve the prediction from $\left(t_0,t_1,t_2,\cdots,t_n\right)$ to  $\left(t_1,t_2,\cdots,t_n,i_1'\right)$). The term $i_1'$ will be regarded as the additional input when predicting the second image token $i_2'$ at the next iteration. The iteration will continue until the last image token $i_m'$ is obtained. Once the autoregression step is finished, we feed the predicted image tokens $\boldsymbol{i'}=\left(i_1',i_2',\cdots,i_m'\right)$ to a pre-trained image decoder (see Section \ref{section:vqvae} for details) to synthesize realistic remote sensing images.

\begin{figure*}
  \centering
  \includegraphics[width=\linewidth]{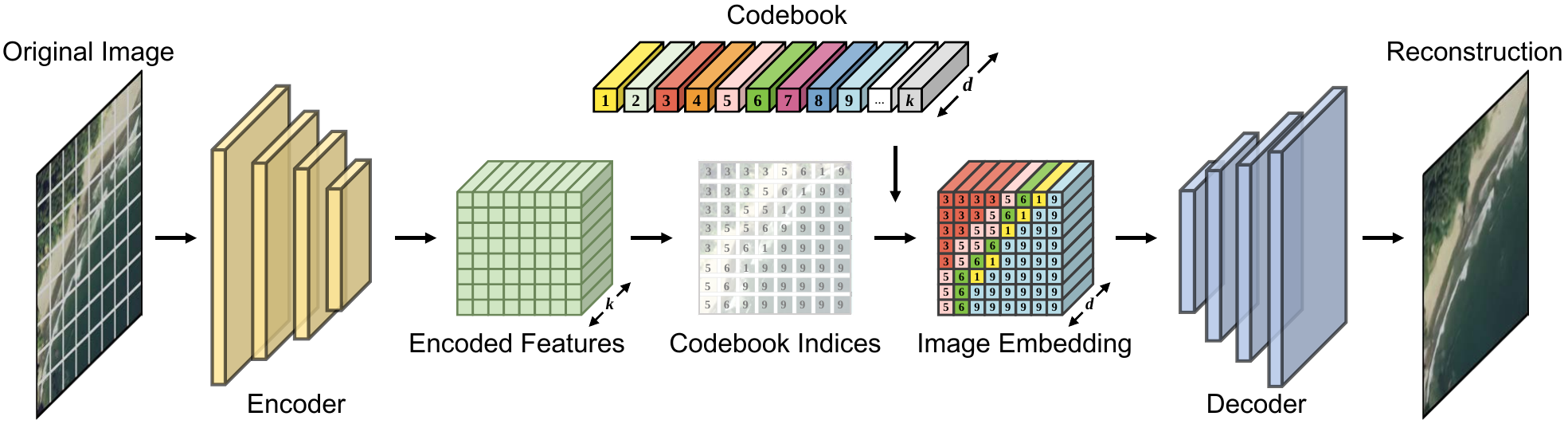}
  \caption{Illustration of the vector quantized variational autoencoder (VQVAE), which consists of an encoder network, a decoder network, and a codebook with learnable parameters. Given an input image, the encoder aims to tokenize the encoded features and obtain the codebook index for each pixel in the encoded features. Once the codebook indices are obtained, we directly copy the corresponding codeword in the codebook to construct the image embedding. Finally, the image embedding is fed into the decoder network to obtain the reconstruction image.}
\label{fig:vqvae}
\end{figure*}

\begin{figure}
  \centering
  \includegraphics[width=\linewidth]{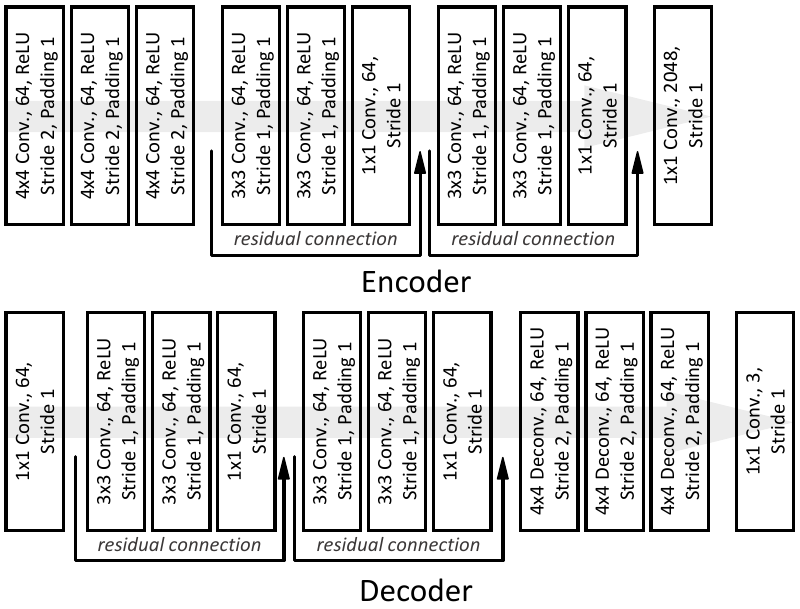}
  \caption{The detailed architectures of the image encoder and decoder networks used in this study.}
\label{fig:architecture}
\end{figure}

\subsection{Image Encoder and Decoder\label{section:vqvae}}
The capability of the image encoder and decoder is significant to the quality of the synthesized images. Inspired by the work in \cite{van2017neural,esser2021taming}, we consider two state-of-the-art image generation models, including the vector quantized variational autoencoder (VQVAE) and the vector quantized generative adversarial network (VQGAN), to learn the image tokenizer in this study. Since the key idea of both VQVAE and VQGAN is to discretize the encoded features, and VQGAN can be regarded as an adversarially trained version of VQVAE, we focus on a detailed introduction of VQVAE in this subsection for the sake of simplicity. Readers can refer to \cite{esser2021taming} for more detailed explanations about the implementation of VQGAN.

As shown in Fig. \ref{fig:vqvae}, there are three main components in the VQVAE: An encoder network, a decoder network, and a codebook with learnable parameters.  Formally, let $Enc\left(\cdot\right)$ and $Dec\left(\cdot\right)$ denote the output of the encoder and decoder networks, respectively (see Fig. \ref{fig:architecture} for detailed architectures of the encoder and decoder networks). Then, we define the codebook $\boldsymbol{c}=\{c_i\}_{i=1}^{k}$, where $c_i\in \mathbb{R}^d$ is the $i$th codeword in $\boldsymbol{c}$, and $k$ and $d$ denote the number of codewords and the feature dimension, respectively. Given an input RGB image $x\in \mathbb{R}^{h\times w\times 3}$, where $h$ and $w$ are the height and width of the image, respectively, the encoded features can thereby be represented as $Enc\left(x\right)\in \mathbb{R}^{\frac{h}{8}\times \frac{w}{8}\times k}$ (the encoder network will downsample the input image with a downsampling rate of $8$). The goal of the encoder network is to tokenize $Enc\left(x\right)$ by predicting the nearest codeword for each pixel in $Enc\left(x\right)$. The codebook indices ${\rm ind}\in \mathbb{R}^{\frac{h}{8}\times \frac{w}{8}}$ can be obtained by
\begin{equation}
{\rm ind}=\arg\max Enc\left(x\right).
\label{eq:indices}
\end{equation}
Note that the codebook indices ${\rm ind}$ in \eqref{eq:indices} is a two-dimensional matrix. To obtain the corresponding image tokens $\boldsymbol{i}$, we can simply flatten ${\rm ind}$ into the vector form.

With the predicted codebook indices ${\rm ind}$, we can then construct the image embedding features ${\rm emb}_{enc}\in \mathbb{R}^{\frac{h}{8}\times \frac{w}{8}\times d}$ by copying the corresponding codewords in the codebook $\boldsymbol{c}$. The reconstructed image $x'\in \mathbb{R}^{h\times w\times 3}$ can be obtained by feeding ${\rm emb}_{enc}$ to the decoder network: $x'=Dec\left({\rm emb}_{enc}\right)$. The whole network is optimized by minimizing the reconstruction loss $\mathcal{L}_{rec}$, which is defined as the mean squared error between the reconstructed image and the original image:
\begin{equation}
\mathcal{L}_{rec}=\frac{1}{hw}\sum_{i=1}^{h}\sum_{j=1}^{w} \|x'_{\left(i,j\right)}-x_{\left(i,j\right)}\|^2.
\label{eq:mse}
\end{equation}

Once the training of VQVAE is finished, the weight parameters in the encoder network, the decoder network, and the codebook will be fixed in the training of the Txt2Img-MHN.

\subsection{Hopfield Layer\label{section:prototype}}
Due to the huge modality gap between text descriptions and remote sensing images, directly learning concrete but highly diverse semantic representations is very challenging. To apply the coarse-to-fine learning strategy, we propose conducting prototype learning in the text-image joint feature space. Specifically, we propose the prototype learning block, which consists of a Hopfield layer and a self-attention layer. The Hopfield layer is originally proposed in the modern Hopfield network \cite{Ramsauer:21}, which can learn static queries or stored patterns. Given that there are abundant high-level semantic concepts contained in the training text-image pairs, our goal is to make the network automatically exploit the clusters of these concepts, and learn the most representative prototypes from the input text-image embeddings hierarchically. These learned prototypes can then be regarded as codewords to represent more complex semantic concepts.

\begin{figure}
  \centering
  \includegraphics[width=\linewidth]{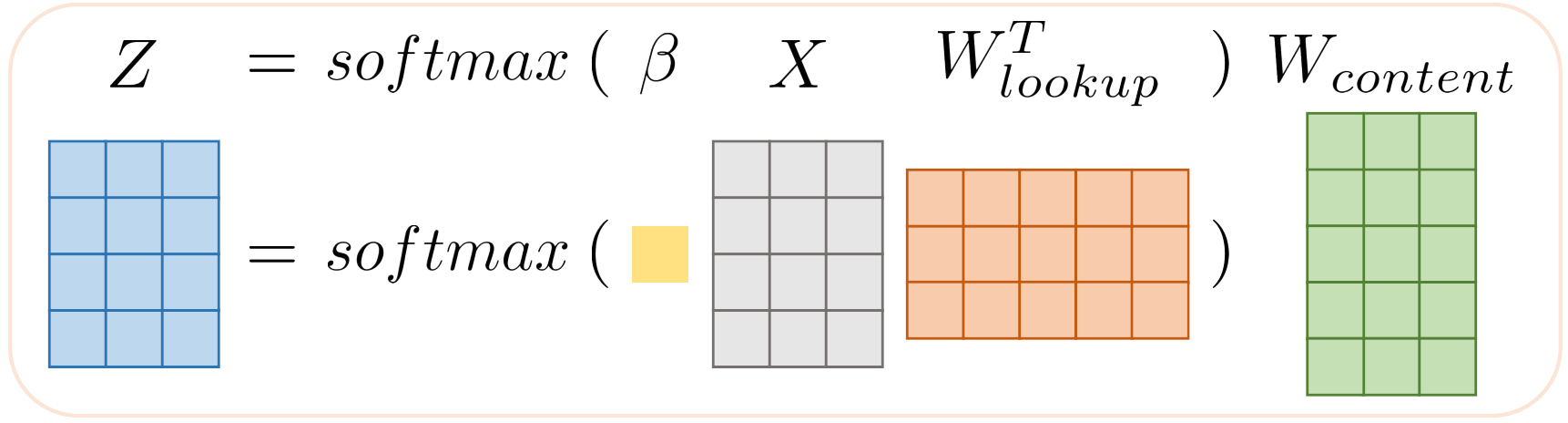}
  \caption{Illustration of the Hopfield layer, which aims to learn the prototype lookup matrix $W_{lookup}^T$ and the prototype content matrix $W_{content}^T$ given the input text-image embedding features $X$.}
\label{fig:hopfield_layer}
\end{figure}

The detailed architecture of the Hopfield layer used in this study is shown in Fig. \ref{fig:hopfield_layer}. Formally, let $X\in \mathbb{R}^{(n+m)\times d_{emb}}$ denote the input text-image embedding feature $X$. The prototype lookup matrix and prototype content matrix are then defined as $W_{lookup}\in \mathbb{R}^{n_{pro}\times d_{emb}}$ and $W_{content}\in \mathbb{R}^{n_{pro}\times d_{emb}}$, respectively, where $n_{pro}$ is the number of prototypes. The output feature $Z\in \mathbb{R}^{(n+m)\times d_{emb}}$ can thereby be formulated as:
\begin{equation}
Z=softmax\left(\beta X W_{lookup}^T\right)W_{content},
\label{eq:pl}
\end{equation}
where $softmax\left(\cdot\right)$ denotes the softmax function, and $\beta=1/\sqrt{d_{emb}}$ is a scaling scalar.

\begin{figure}
  \centering
  \includegraphics[width=\linewidth]{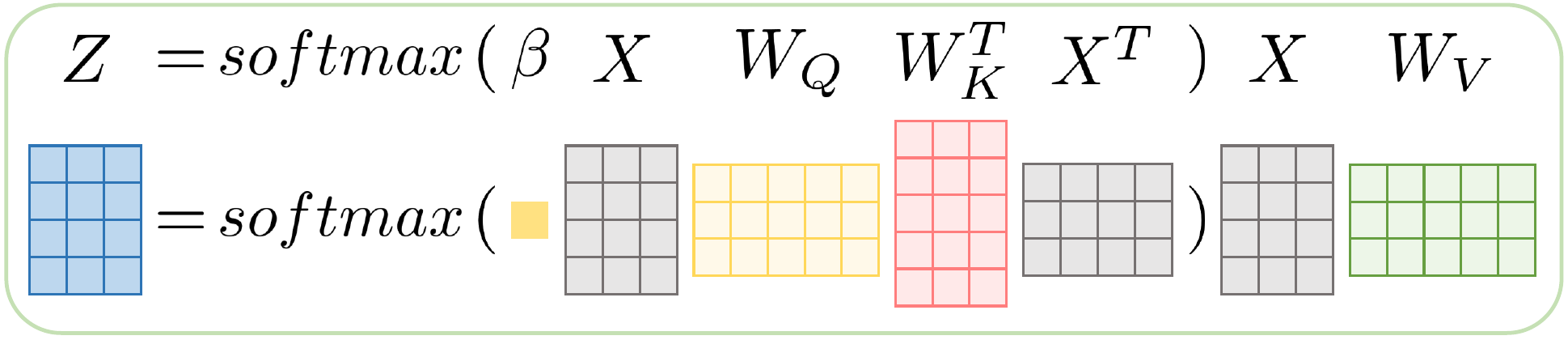}
  \caption{Illustration of the self-attention layer, which aims to strengthen the interaction of the information among different tokens in the input text-image embedding feature $X$.}
\label{fig:sa_layer}
\end{figure}

With \eqref{eq:pl}, the Hopfield layer can achieve the prototype learning from two perspectives: 1) learning the most representative prototypes and saving the stored patterns in $W_{content}$, where each row in $W_{content}$ corresponds to a learned text-image prototype; and 2) learning to represent the input text-image embedding features $X$ with the learned prototypes using the prototype lookup matrix $W_{lookup}$. Thus, the first term in \eqref{eq:pl} (i.e., $softmax\left(\beta X W_{lookup}^T\right)$) can be interpreted as the probability or portion of each prototype in the mapping.

\subsection{Self-Attention Layer}
Self-attention learning is a commonly used technique in machine translation \cite{shaw2018self} and image interpretation \cite{xu2021self}. Inspired by the work in \cite{vaswani2017attention}, we propose to use the self-attention layer to further strengthen the inner information interaction among different tokens in the input text-image embedding feature $X$, as shown in Fig.~\ref{fig:sa_layer}. Formally, let $\{W_Q, W_K, W_V\} \in \mathbb{R}^{d_{emb}\times d_{emb}}$ denote the query, key, and value transformation matrices, respectively. Then, the output feature $Z\in \mathbb{R}^{(n+m)\times d_{emb}}$ can thereby be formulated as:
\begin{equation}
Z=softmax\left(\beta X W_{Q}W_{K}^TX^T\right)XW_{V},
\label{eq:attention}
\end{equation}
where the first term (i.e., $softmax\left(\beta X W_{Q}W_{K}^TX^T\right)$) measures the impact of each position on all other positions in the text-image embedding.

\subsection{Optimization}
Recall that our goal is to predict the image tokens that can be used to synthesize realistic images with the decoder network trained in Section \ref{section:vqvae}. To achieve this goal, we feed the features learned from the cascaded prototype learning blocks into a new Hopfield layer with the prototype content matrix $W_{content}'\in \mathbb{R}^{n_{pro}\times k}$, where $k$ is the number of codewords defined in Section \ref{section:vqvae}. Unlike previous prototype learning layers, each row in $W_{content}'$ corresponds to a learned high-level semantic prototype that may contain multiple codewords in the codebook $\boldsymbol c$ with varying proportions. Then, the input text-image embedding feature can be represented with the learned prototypes and transformed into the $k$-dimensional image token space. The output feature $Z'=softmax\left(\beta X W_{lookup}^T\right)W_{content}'\in \mathbb{R}^{\left(n+m\right)\times k}$ can be regarded as the predicted logits. Note that the first $n$ rows in $Z'$ correspond to the $n$ text tokens. Thus, the predicted logit for the first image token is at the $\left(n+1\right)$th row in $Z'$. Let $p_i$ denotes the logits for the $i$th image token. The cross-entropy loss can be defined as:
 \begin{equation}
\mathcal{L}_{ce}=-\sum_{i=1}^{m}\sum_{j=1}^{k}y_{i_j}\log softmax(p_i)_j,
\label{eq:ce}
\end{equation}
where $y_i$ is the one-hot form of the $i$th input image token $i_i$ (ground truth).

In the test phase, since there are only text descriptions as the input, the predicted image tokens will be obtained auto-regressively \cite{graves2013generating}. Specifically, given the text tokens $\boldsymbol{t}=\left(t_1,t_2,\cdots,t_n\right)$, the network will predict the first image token $i_1'=\arg\max p_1$ at the first iteration. Then, $i_1'$ will be regarded as the additional input when predicting the second image token $i_2'$ at the next iteration. In this way, we can obtain all $m$ predicted image tokens and feed them to the image decoder (pre-trained in Section \ref{section:vqvae}) to synthesize realistic remote sensing images.
\section{Experiments}
\subsection{Experimental Settings and Implementation Details}
\textbf{Dataset:} We use the RSICD dataset\footnote{https://github.com/201528014227051/RSICD\_optimal} \cite{lu2017exploring} to evaluate the proposed method in this study. The RSICD dataset was originally collected for the remote sensing image captioning task. It contains a total of 10921 aerial remote sensing images with various resolutions collected from Google Earth, Baidu Map, MapABC, and Tianditu. Each image has a size of $224\times 224$ and is annotated with 5 text descriptions. There are 30 scene classes in the dataset: airport, bareland, baseball field, beach, bridge, center, church, commercial, dense residential, desert, farmland, forest, industrial, meadow, medium residential, mountain, parking, park, playground, pond, port, railway station, resort, river, school, sparse residential, square, stadium, storage tanks, and viaduct. To ensure fair comparisons, we adopt the same experimental setting as used in \cite{zhao2021text}. Specifically, we use the 8734 text-image pairs from the training split of the RSICD dataset as the training set for each method reported in this study, while the remaining 2187 text-image pairs serve as the test set. For each caption in the test set, we generate 10 images to compute the evaluation metrics. During the experiments, all images are scaled to a resolution of $256 \times 256$. For the comparing methods reported in this study, we use the same hyper-parameter settings as those used in the original paper to train the model on the RSICD dataset.

\begin{figure*}
  \centering
  \includegraphics[width=0.85\linewidth]{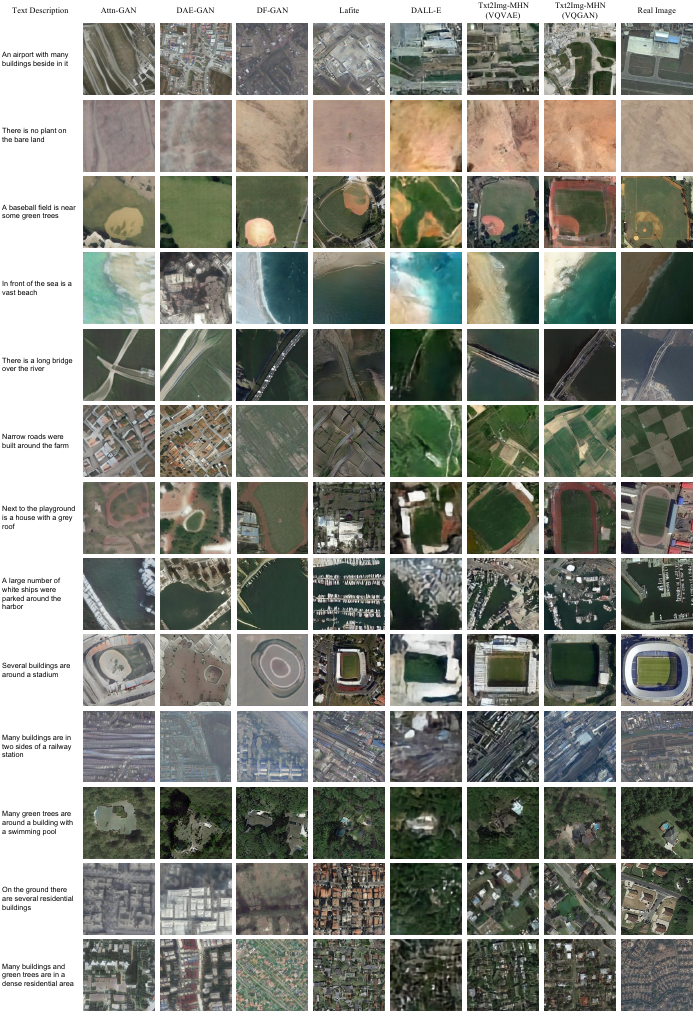}
  \caption{Illustration of the remote sensing images generated based on the text descriptions from the test set, using different text-to-image generation methods.}
\label{fig:gen}
\end{figure*}

\begin{figure}
  \centering
  \includegraphics[width=\linewidth]{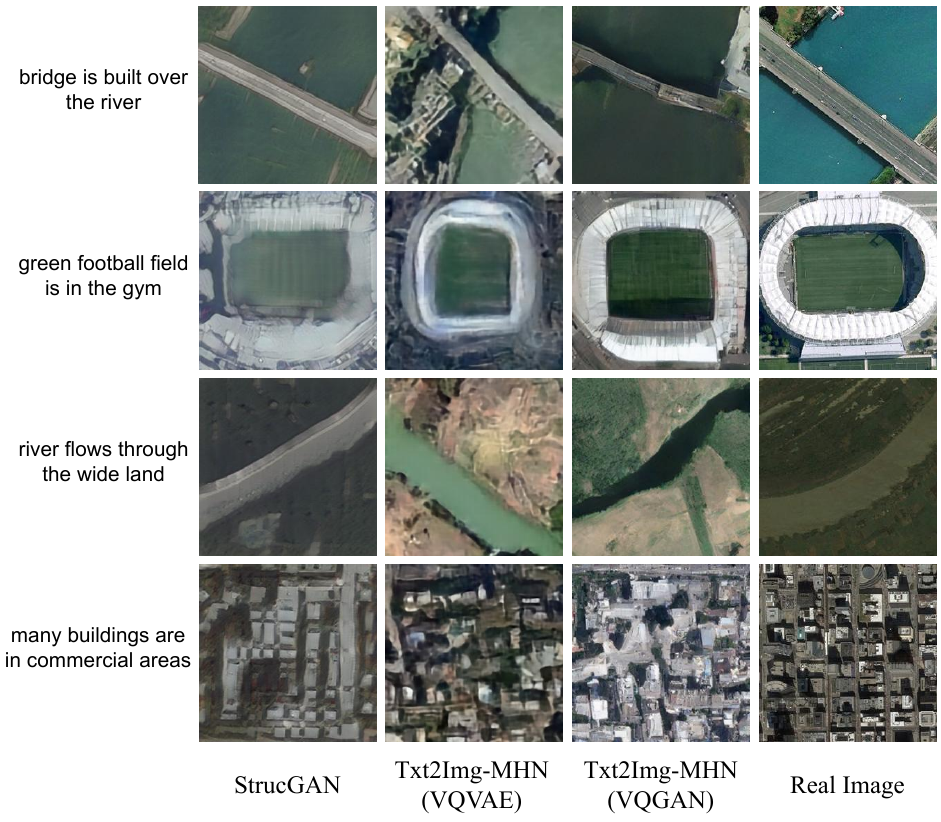}
  \caption{llustration of the synthesized remote sensing images using StrucGAN \cite{zhao2021text} and the proposed Txt2Img-MHN.}
\label{fig:StrucGAN}
\end{figure}

\textbf{Evaluation metrics:} Four evaluation metrics are adopted to evaluate the quality of text-to-image generation, including the Inception Score \cite{salimans2016improved}, Fr\'{e}chet Inception Distance (FID) \cite{heusel2017gans}, CLIP Score \cite{radford2021learning}, and the proposed zero-shot classification overall accuracy (OA). A brief introduction for each evaluation metric is as follows:
\\1) Inception Score: Formally, let $x'$ denote the generated image to be evaluated, and $q\left(l|x'\right)$ is the posterior probability of a label $l$ calculated by the RSICD pre-trained Inception-V3 model \cite{inception} on $x'$. Then, the Inception Score (IS) can be calculated as
\begin{equation}
{\rm IS}=\exp \left(\mathbb{E}_{x'}\left[D_{\rm KL}\left(q\left(l|x'\right)\right\|q\left(l\right))\right]\right),
\label{eq:is}
\end{equation}
where $q\left(l\right)$ is the marginal class distribution, and $D_{\rm KL}\left(\cdot\right)$ denotes the KL-divergence between two probability distributions.
\\2) FID: We use the features obtained from the final average pooling layer in the RSICD pre-trained Inception-V3 model to compute the FID score:
\begin{equation}
{\rm FID}=\|\mu_r-\mu_g\|^2+Tr\left(\Sigma_r+\Sigma_g-2\left(\Sigma_r\Sigma_g\right)^\frac{1}{2}\right),
\label{eq:fid}
\end{equation}
where $\left(\mu_r,\Sigma_r\right)$ and $\left(\mu_g,\Sigma_g\right)$ denote the mean and covariance of the real and generated features, respectively, and $Tr\left(\cdot\right)$ represents the trace linear algebra operation.
\\3) CLIP Score: This metric measures the mean cosine similarity between the text descriptions and the generated images in the feature space of a CLIP model. The CLIP-RSICD-V2\footnote{https://huggingface.co/flax-community/clip-rsicd-v2} model, fine-tuned on the RSICD dataset, is adopted in the evaluation.
\\4) Zero-shot classification OA: It should be noted that both IS and FID were originally proposed to evaluate GAN models, and may not accurately measure the semantic consistency between the generated images and the original samples. CLIP Score only evaluates the similarity between the generated image and the input text description, without taking into account other factors such as visual quality, diversity, and realism of the generated image, as discussed in the previous work \cite{saharia2022photorealistic}. To address these limitations, we further propose to conduct zero-shot classification on the real test set and adopt overall accuracy (OA) as the quantitative evaluation metric for text-to-image generation for the first time. Specifically, for each text-to-image generation method used in this study, we train a classification model (ResNet-18 \cite{resnet}) based on the generated images using the text descriptions on the test set. These trained classification models are then adopted to conduct zero-shot classification on all images in the original test set (the label of these real images can be easily collected, as the images in RSICD are named with the corresponding category, e.g., \texttt{airport\_1.jpg}). Since no test images are used in the training of these classification models, the obtained OA scores can reflect the quality and semantic consistency of the generated images well.

\textbf{Implementation details:} We adopt the Adam optimizer with the initial learning rate of $1e-3$ to train the VQVAE and VQGAN models. The ``exponential'' learning rate decay policy is used, where the initial learning rate is multiplied by $0.996$ at each epoch. The number of total training epochs is set to $1000$ with a batch size of $256$. For the training of the proposed Txt2Img-MHN, we adopt the Adam optimizer with an initial learning rate of $4.5e-3$. The ``exponential'' learning rate decay policy is used here as well, and the learning rate decay parameter is set to $0.999$. We set the maximum length of the input sentence to $n=40$ and the total length of the image tokens to $m=1024$ in this study. The feature dimension of the codeword $d$ and the number of codewords $k$ are set to $512$ and $2048$, respectively. The number of prototypes $n_{pro}$ and the feature dimension in the embedding layers $d_{emb}$ are set to $1000$ and $512$, respectively. The number of prototype learning blocks is fixed to $N=10$. See Section~\ref{section:abl} for detailed analyses of the most important parameters. To conduct zero-shot classification, we generate $10$ synthetic images for each text-image pair in the test set, resulting in a total of $20540$ synthetic images (i.e., $2054\times 10$, excluding the $133$ images without category labels from the original test set). Subsequently, we train a ResNet-18 model for each text-to-image generation method using their respective synthetic training sets. The training process employs the Adam optimizer with a learning rate of $1e-4$. The number of total training epochs is set to $5$ with a batch size of $64$. Once the training is completed, these ResNet-18 models are then utilized to conduct zero-shot classification on the $2054$ real images in the original test set, and the corresponding OA scores are obtained.
The experiments in this paper are conducted in PyTorch using 8 NVIDIA Tesla A100 GPUs.

\subsection{Qualitative Results}

In this subsection, we evaluate the qualitative performance of the proposed method along with several recent state-of-the-art approaches. A brief introduction to these methods is given below.

\begin{itemize}
\item Attn-GAN (Attentional GAN, CVPR 2018)\footnote{https://github.com/taoxugit/AttnGAN}: Fine-grained text-to-image generation based on GAN with attention learning and multi-stage refinement \cite{xu2018attngan}.
\item DAE-GAN (Dynamic Aspect-awarE GAN, ICCV 2021)\footnote{https://github.com/hiarsal/DAE-GAN}: Aspect-level text-to-image synthesis with attended global refinement and aspect-aware local refinement \cite{ruan2021dae}.
\item StrucGAN (Structured GAN, IEEE GRSL 2021): A multi-stage framework that synthesizes remote sensing images with structured information \cite{zhao2021text}.
\item DF-GAN (Deep Fusion GAN, CVPR 2022)\footnote{https://github.com/tobran/DF-GAN}: GAN-based text-to-image generation using the deep text-image fusion block and a target-aware discriminator \cite{tao2022df}.
\item Lafite (CVPR 2022)\footnote{https://github.com/drboog/Lafite}: Language-free text-to-image generation with the pre-trained CLIP (Contrastive Language-Image Pre-training) model \cite{zhou2022towards}.
\item DALL-E (ICML 2021)\footnote{https://github.com/lucidrains/DALLE-pytorch}: Zero-shot text-to-image generation using the modified GPT-3 (generative pre-trained transformer-3) model \cite{ramesh2021zero}.
\item Txt2Img-MHN (VQVAE): The proposed text-to-image modern Hopfield network with image encoder and decoder from VQVAE.
\item Txt2Img-MHN (VQGAN): The proposed text-to-image modern Hopfield network with image encoder and decoder from VQGAN.
\end{itemize}

In general, the aforementioned GAN-based methods are trained in an adversarial manner, where the generator network takes a textual description as input and synthesizes an image, while the discriminator network evaluates the realism of the generated image. In contrast, Transformer-based methods like DALL-E use a Transformer network to simultaneously learn feature embeddings for both the image and text tokens, enabling text-to-image generation in an auto-regressive manner without adversarial training.

Some examples of the generated remote sensing images based on the text descriptions from the test set using different text-to-image generation methods are presented in Fig. \ref{fig:gen}. Due to the complex spatial distribution of different ground objects, generating photo-realistic and semantic-consistent remote sensing images is very challenging. While state-of-the-art GAN-based methods like DF-GAN can synthesize the visual appearance style of the remote sensing images well, they may fail to accurately construct the detailed shape and boundary information for complex ground targets like the playground (7th row in Fig. \ref{fig:gen}) and the stadium (9th row in Fig. \ref{fig:gen}). By contrast, the proposed Txt2Img-MHN can yield more realistic results in these challenging scenarios. Moreover, we find that Txt2Img-MHN can better learn the concept of quantity. Take the last three rows in Fig. \ref{fig:gen} as an example. While all these three input text descriptions are related to the building object, the quantity of the buildings in the text descriptions vary from ``\texttt{a building},'' or ``\texttt{several buildings},'' to ``\texttt{many buildings}.'' Although most of the methods can generate relatively good results in the first scenario, accurately understanding the concept of ``\texttt{several}'' and ``\texttt{many}'' is more difficult. This phenomenon is especially obvious in the result of Lafite, where it incorrectly generates a lot of dense residential buildings in the case of the input text descriptions of ``\texttt{several buildings}.'' By contrast, the results of Txt2Img-MHN are more semantically-consistent with the input text descriptions.

\begin{figure}
  \centering
  \includegraphics[width=\linewidth]{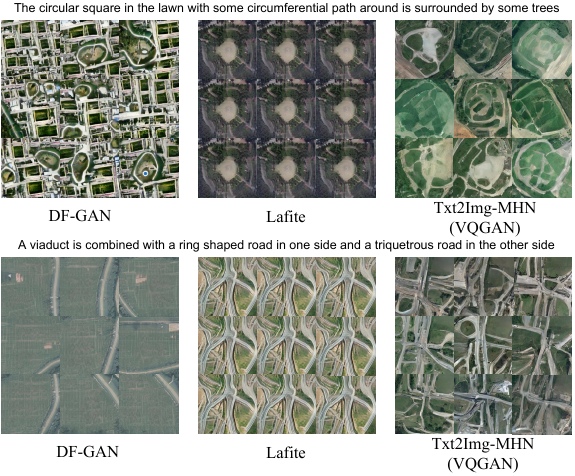}
  \caption{Illustration of the diversity of the generated images. For each input text description from the test set, we randomly run different text-to-image generation methods with 9 repetitions. Note: we do not manually ``cherry pick'' any of the generated images.}
\label{fig:nopick}
\end{figure}

We compared our proposed Txt2Img-MHN with the StrucGAN model proposed in \cite{zhao2021text}. As the source code for StrucGAN was not available, we used the same text descriptions as in \cite{zhao2021text} to synthesize remote sensing images with Txt2Img-MHN, while replicating StrucGAN's results from the original paper. The results are shown in Fig.~\ref{fig:StrucGAN}. Additionally, StrucGAN's output ground objects appeared greyish, as can be seen in the last two rows of Fig.~\ref{fig:StrucGAN}, resulting in less realistic outcomes. In contrast, Txt2Img-MHN better simulated the color and texture information of various ground objects.

\begin{figure*}
  \centering
  \includegraphics[width=\linewidth]{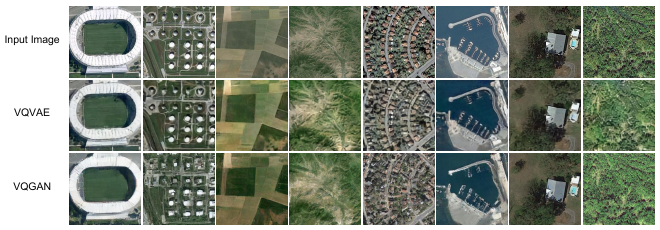}
  \caption{Image reconstruction using VQVAE and VQGAN. 1st row: Input remote sensing images from the test set. 2nd and 3rd rows: Reconstructed images from VQVAE and VQGAN, respectively.}
\label{fig:recons}
\end{figure*}

\begin{table*}
\caption{Quantitative Results on the RSICD Test Set. We Report the Inception Score, the FID Score, the CLIP Score, and the Overall Accuracy (\%) in the Zero-Shot Classification}
\centering
\resizebox{\textwidth}{!}{%
\begin{tabular}{cccccc}
\toprule
Method & Approach & Inception Score $\uparrow$ & FID Score $\downarrow$ & CLIP Score ($\times$100) $\uparrow$& Zero-Shot Classification OA $\uparrow$\\
\midrule
Attn-GAN \cite{xu2018attngan} & \multirow{5}{*}{GAN-based} &\textbf{11.71} &95.81 &20.19&32.46  \\
DAE-GAN \cite{ruan2021dae} &&7.71 &93.15 &19.69 & 29.74  \\
StrucGAN \cite{zhao2021text} & &5.84$^*$ &-- &--&--  \\
DF-GAN \cite{tao2022df} &&9.51  &109.41 &19.76  &51.99  \\
Lafite \cite{zhou2022towards} &&10.70  &\textbf{74.11}& \textbf{22.52}& 49.37  \\
\midrule
DALL-E \cite{ramesh2021zero} & \multirow{3}{*}{Transformer-based} &2.59 &191.93&20.13&28.59  \\
Txt2Img-MHN (VQVAE) &&3.51  &175.36 & 21.35 &41.46  \\
Txt2Img-MHN (VQGAN) && 5.99 & 102.44& 20.27&\textbf{65.72}\\
\bottomrule
\end{tabular}}
\label{tab:result}
\vspace{2pt}
\leftline{\scriptsize Note: Best results are highlighted in \textbf{bold}. $^*$The Inception score of StrucGAN is replicated from the original paper.}
\end{table*}

One characteristic of the proposed Txt2Img-MHN is the diversity of the generated images. Fig. \ref{fig:nopick} illustrates some examples of the generated images from DF-GAN, Lafite, and the proposed Txt2Img-MHN (with VQGAN). For each input text description from the test set, we randomly run these methods 9 times and directly visualize all of the generated images without any manual cherry picking. It can be observed that the results of Lafite are highly static and unvaried among different repetitions. DF-GAN can synthesize more diverse images, but may not maintain high semantic consistency with the input text descriptions. By contrast, the proposed Txt2Img-MHN can achieve diverse generations given the same text descriptions, while preserving relatively high generation quality and semantic consistency. This feature can benefit real application scenarios like urban planning, where the synthesized remote sensing images for squares or viaducts with high diversity may inspire urban planning researchers.

To better understand how VQVAE and VQGAN would influence the performance of the proposed method on the text-to-image generation task in a remote sensing scenario, we further evaluate the reconstruction ability of VQVAE and VQGAN. As shown in Fig. \ref{fig:recons}, given an input remote sensing image from the test set, we visualize the reconstruction results from VQVAE and VQGAN. Compared to VQVAE, VQGAN can yield more detailed reconstructions and preserve better texture information owing to the adversarial training strategy. This phenomenon is also consistent with the results in Fig.~\ref{fig:gen}, where Txt2Img-MHN (VQGAN) can generally synthesize images with a higher spatial resolution. However, there exists the risk that the shapes of different ground objects would be changed in VQGAN's reconstruction. For example, the round storage tanks in the second column of Fig. \ref{fig:recons} are reconstructed to be rectangular. By contrast, the results of VQVAE are more blurry but can reconstruct the shape information more accurately.

\subsection{Quantitative Evaluation with Zero-Shot Classification}
While it is usually intuitive to evaluate the quality of the synthesized images from their visual appearance to human observers, conducting accurate quantitative evaluation remains a challenging task considering the ambiguity and complexity of semantic information contained in the image. Currently, the most commonly adopted metrics for image generation are the Inception Score and the FID Score. Despite their simplicity, these two metrics were originally used to evaluate the quality of the GAN model and focus more on either the visual style similarity or the feature distance between synthesized and real data. Thus, they may not accurately reflect the semantic consistency of the generated images. CLIP Score can provide an estimate of the image-text alignment but it only evaluates the similarity between the generated image and the input text description, without taking into account other factors such as visual quality, diversity, and realism of the generated image \cite{saharia2022photorealistic}. To address these limitations, we propose to conduct zero-shot classification on the real test images with the classification model trained on synthesized data. Specifically, for each text-to-image generation method used in this study, we train a ResNet-18 model with the synthesized images generated from the text descriptions from the test set in RSICD. Then, we evaluate the zero-shot classification accuracy of each ResNet-18 model on the real test images in RSICD and report the OA performance. Since no real data are used in the training of these classification models, the obtained OA in the zero-shot classification can measure the quality and semantic consistency of the generated images well.

\begin{table}
\caption{Overall Accuracy (\%) of the Zero-Shot Classification with Different Values of $n_{pro}$}
\centering
\begin{tabular*}{\linewidth}{@{\extracolsep{\fill}}c|ccccc}
\toprule
$n_{pro}$&10&100&500&1000&1500\\
\midrule
Txt2Img-MHN (VQVAE) &29.69&35.19&39.92&\textbf{41.46}&40.21\\
Txt2Img-MHN (VQGAN) &38.74&54.33&\textbf{66.38}&65.72&66.13\\
\bottomrule
\end{tabular*}
\\
\vspace{2pt}
\leftline{\scriptsize Note: Best results are highlighted in \textbf{bold}.}
\label{tab:prototype}
\end{table}

\begin{table}
\caption{Overall Accuracy (\%) of the Zero-Shot Classification with Different Values of $d_{emb}$}
\centering
\begin{tabular*}{\linewidth}{@{\extracolsep{\fill}}c|ccccc}
\toprule
$d_{emb}$&64&128&256&512&1024\\
\midrule
Txt2Img-MHN (VQVAE) &30.96&33.00&34.85&\textbf{41.46}&38.30\\
Txt2Img-MHN (VQGAN) &52.14&55.64&60.07&\textbf{65.72}&57.93\\
\bottomrule
\end{tabular*}
\\
\vspace{2pt}
\leftline{\scriptsize Note: Best results are highlighted in \textbf{bold}.}
\label{tab:d}
\end{table}

\begin{table}[!t]
\caption{Overall Accuracy (\%) of the Zero-Shot Classification with Different Values of $N$}
\centering
\begin{tabular*}{\linewidth}{@{\extracolsep{\fill}}c|ccccc}
\toprule
$N$&1&2&5&10&15\\
\midrule
Txt2Img-MHN (VQVAE) &20.74&26.43&37.34&\textbf{41.46}&39.82\\
Txt2Img-MHN (VQGAN) &46.88&50.92&59.73&65.72&\textbf{65.92}\\
\bottomrule
\end{tabular*}
\\
\vspace{2pt}
\leftline{\scriptsize Note: Best results are highlighted in \textbf{bold}.}
\label{tab:N}
\end{table}

Table \ref{tab:result} presents detailed results for the zero-shot classification OA, Inception Score, FID Score, and CLIP Score for each method used in this study. Although evaluating the quality of generated images visually is intuitive, it is challenging to perform accurate quantitative evaluation due to the ambiguity and complexity of semantic information in the image. Lafite has an advantage in the CLIP Score metric since it directly applies a contrastive loss for the generator using the pre-trained CLIP model to constrain the consistency of the generated image and its corresponding text feature, resulting in the highest score of 22.52. The proposed Txt2Img-MHN (VQVAE) obtains a good result on this metric, ranking second with a score of 21.35. Interestingly, GAN-based methods generally perform much better on the Inception Score and FID Score than Transformer-based methods, as observed in a previous study \cite{ramesh2021zero}. For instance, Attn-GAN achieves the highest Inception Score of $11.71$, whereas the proposed Txt2Img-MHN (VQGAN) only yields an Inception Score of 5.99. However, a high Inception Score may not always lead to a high zero-shot classification OA. For example, Attn-GAN can only achieve an OA of around 32\%, despite its high Inception Score. On the other hand, the proposed Txt2Img-MHN (VQGAN) achieves an OA of over 65\%, outperforming all comparable methods by a significant margin. This result demonstrates the great potential for data augmentation based on the generated remote sensing images. As remote sensing tasks usually require higher levels of realism and semantic consistency of the synthesized data, we believe that the OA in zero-shot classification can serve as a better criterion for evaluating the text-to-image generation task in the remote sensing scenario.

\subsection{Ablation Study}\label{section:abl}

\begin{table}
\centering
\caption{Overall Accuracy (\%) of the Zero-Shot Classification of the Proposed Txt2Img-MHN (VQGAN) Model with Different Architectures or Strategies}
\label{tab:architecture}
\resizebox{\linewidth}{!}{%
\begin{tabular}{cc}
\toprule
& Zero-shot classification OA \\
\hline
Prototype learning block (Fig.~\ref{fig:ablation}(a))& \textbf{65.72} \\
Prototype learning block w/o Hopfield layer (Fig.~\ref{fig:ablation}(b))& 57.10 \\
Prototype learning block w/o self-attention layer (Fig.~\ref{fig:ablation}(c))& 36.50 \\
Prototype learning block w/ an inverse order (Fig.~\ref{fig:ablation}(d))& 61.31 \\
Prototype learning block w/ parameter sharing (Fig.~\ref{fig:ablation}(e))& 55.30 \\
\bottomrule
\end{tabular}
}
\\
\vspace{2pt}
\leftline{\scriptsize Note: Best results are highlighted in \textbf{bold}.}
\end{table}

In this subsection, we provide a detailed analysis of how varying the hyperparameters in the proposed Txt2Img-MHN affects its performance. Specifically, we examine the impact of the number of prototypes ($n_{pro}$), the feature dimension in the embedding layers ($d_{emb}$), and the number of prototype learning blocks ($N$) on the model's performance. Our experiments show that using excessively small values for these hyperparameters can limit the prototype learning ability and result in relatively low overall accuracy in zero-shot classification (see Tables~\ref{tab:prototype} to \ref{tab:N}). To strike the right balance between performance and computational efficiency, we empirically set $n_{pro}=1000$, $d_{emb}=512$, and $N=10$ for this study. These values have been found to yield good results in our experiments and can serve as a guideline for future studies in this area.

\begin{figure}
  \centering
  \includegraphics[width=\linewidth]{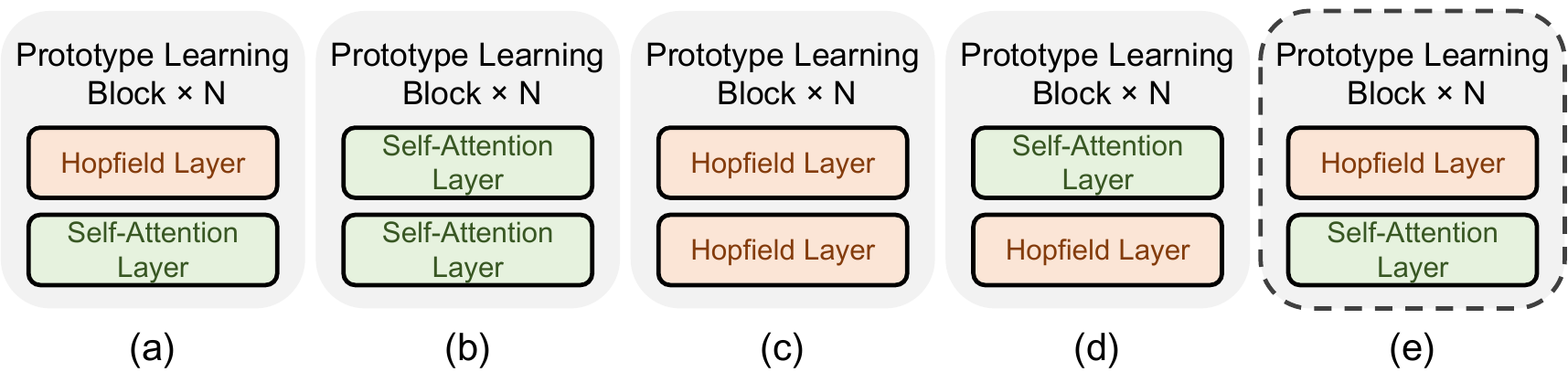}
  \caption{Various variants of the proposed prototype learning block. (a) The original architecture. (b) Prototype learning block without Hopfeld layer. (c) Prototype learning block without self-attention layer. (d) Prototype learning block with an inverse order. (e) Prototype learning block with parameter sharing.}
\label{fig:ablation}
\end{figure}

Table~\ref{tab:architecture} presents a detailed ablation study to assess the impact of the Hopfield layer and self-attention layer. An illustration of various variants of the proposed prototype learning block used in the ablation study is presented in Fig.~\ref{fig:ablation}. The results show that both layers significantly contribute to the performance, and removing either layer from the prototype learning block leads to decreased accuracy, with a greater impact observed for the self-attention layer. Moreover, reversing the order of the two layers (i.e., placing the self-attention layer before the Hopfield layer) also results in reduced zero-shot classification accuracy. Notably, we observe that parameter sharing has a detrimental effect on the performance as it limits the learning capacity of our proposed method.

\begin{figure}
  \centering
  \includegraphics[width=\linewidth]{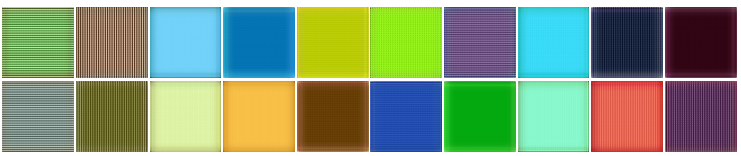}
  \caption{Visualization of the top 20 tokens learned by the last Hopfield layer in the proposed method.}
\label{fig:prototypes}
\end{figure}

To gain a more comprehensive understanding of the Hopfield layer's learned features, we present a visualization of the top 20 tokens learned by the last Hopfield layer. This involves utilizing the learned weights in the prototype content matrix $W_{content}'$ and selecting the tokens with the highest global activation intensity, which indicates the most frequently occurring features in the generated images. As a reminder, each row in $W_{content}'$ corresponds to a learned prototype with the activation intensity for those $k$ codewords. Specifically, we first sum up $W_{content}'$ in the prototype dimension to obtain the global activation intensity for all $k$ codewords: $GAI^{\left(j\right)}=\sum_{i=1}^{n_{pro}} W_{content}'^{\left(i,j\right)}$, where $GAI^{\left(j\right)}$ denotes the global activation intensity for the $j$th codeword, with $j$ ranging from 1 to $k$. We then select the top 20 element indices based on the GAI value from largest to smallest. For each of these indices, we replicate its corresponding codeword from the codebook $\boldsymbol{c}$ in VQVAE $wh/64$ times to construct the image embedding features ${\rm emb'}_{enc}\in \mathbb{R}^{\frac{h}{8}\times \frac{w}{8}\times d}$. The visualized token image is then obtained by inputting ${\rm emb'}_{enc}$ to the decoder network: $Dec\left({\rm emb'}_{enc}\right)$. The resulting visualization in Fig.~\ref{fig:prototypes} reveals that the top 20 tokens learned by the last Hopfield layer correspond to the basic components used for image generation, such as various colors and texture patterns. Moreover, some of these tokens have high similarity to specific ground objects, such as meadows or deserts.

\begin{figure}
  \centering
  \includegraphics[width=\linewidth]{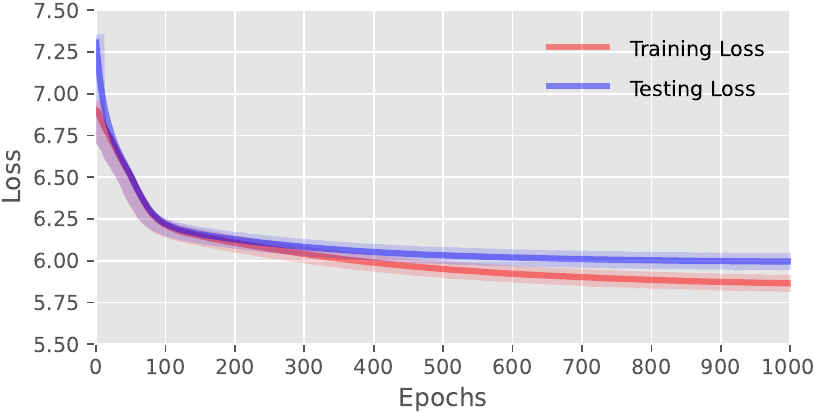}
  \caption{Training and testing losses of the proposed Txt2Img-MHN (VQGAN) at various epochs.}
\label{fig:loss}
\end{figure}

Fig.~\ref{fig:loss} presents the training and testing losses of the proposed Txt2Img-MHN (VQGAN) at various epochs. It can be observed that as the number of epochs increases, both the training loss and testing loss steadily decrease, eventually converging and stabilizing.

\section{Conclusions and Discussions}
In this study, we propose a novel text-to-image modern Hopfield network (Txt2Img-MHN) to generate photo-realistic and semantic-consistent remote sensing images using text descriptions. Instead of learning concrete but highly diverse text-image joint representation directly, Txt2Img-MHN aims to hierarchically learn the most representative prototypes from the input text-image embeddings, which can be further utilized to represent more complex semantics. Additionally, we comprehensively compare and analyze the performance of the vector quantized variational autoencoder (VQVAE) and vector quantized generative adversarial network (VQGAN) for text-to-image generation in the remote sensing scenario. The image reconstruction results show that VQVAE can better preserve the shape information of ground objects, while VQGAN can provide more detailed texture and boundary information. Since traditional evaluation metrics like the Inception Score and FID Score may not accurately measure the realism and semantic consistency of the generated images, we propose a new metric to evaluate the performance of the text-to-image generation task. For this purpose, we further conduct zero-shot classification on the real images with the classification model trained on the synthesized data. Despite its simplicity, we find that the OA in the zero-shot classification may serve as a better criterion for measuring the quality of the image generation. Extensive experiments on the benchmark remote sensing text-image dataset RSICD demonstrate that the proposed method can generate more realistic remote sensing images than the existing approaches.

One of the main challenges in remote sensing image generation from text is the limited text-image pairs in the training set. While high-quality text descriptions for remote sensing data are usually difficult to collect, there is a large quantity of unlabeled high-resolution remote sensing images. Thus, making use of the abundant semantic information in these unlabeled images to improve the performance of text-to-image generation deserves further study. We will try to explore this topic in our future work.

\section*{Acknowledgment}
The authors would like to thank Bernhard Schäfl, Johannes Lehner, Andreas Fürst, Daniel Springer, Shizhen Chang, and Angela Bitto-Nemling for their valuable comments and discussions, and the Institute of Advanced Research in Artificial Intelligence (IARAI) for its support.

\bibliographystyle{IEEEtran}

\bibliography{MHN}

\begin{thebibliography}{10}
\providecommand{\url}[1]{#1}
\csname url@samestyle\endcsname
\providecommand{\newblock}{\relax}
\providecommand{\bibinfo}[2]{#2}
\providecommand{\BIBentrySTDinterwordspacing}{\spaceskip=0pt\relax}
\providecommand{\BIBentryALTinterwordstretchfactor}{4}
\providecommand{\BIBentryALTinterwordspacing}{\spaceskip=\fontdimen2\font plus
\BIBentryALTinterwordstretchfactor\fontdimen3\font minus
  \fontdimen4\font\relax}
\providecommand{\BIBforeignlanguage}[2]{{%
\expandafter\ifx\csname l@#1\endcsname\relax
\typeout{** WARNING: IEEEtran.bst: No hyphenation pattern has been}%
\typeout{** loaded for the language `#1'. Using the pattern for}%
\typeout{** the default language instead.}%
\else
\language=\csname l@#1\endcsname
\fi
#2}}
\providecommand{\BIBdecl}{\relax}
\BIBdecl

\bibitem{ghamisi2017advanced}
P.~Ghamisi, J.~Plaza, Y.~Chen, J.~Li, and A.~J. Plaza, ``Advanced spectral
  classifiers for hyperspectral images: A review,'' \emph{IEEE Geosci. Remote
  Sens. Mag.}, vol.~5, no.~1, pp. 8--32, 2017.

\bibitem{uaers}
Y.~Xu and P.~Ghamisi, ``Universal adversarial examples in remote sensing:
  Methodology and benchmark,'' \emph{IEEE Trans. Geos. Remote Sens.}, vol.~60,
  pp. 1--15, 2022.

\bibitem{zhang2022artificial}
L.~Zhang and L.~Zhang, ``Artificial intelligence for remote sensing data
  analysis: A review of challenges and opportunities,'' \emph{IEEE Geosci.
  Remote Sens. Mag.}, vol.~10, no.~2, pp. 270--294, 2022.

\bibitem{cheng2020remote}
G.~Cheng, X.~Xie, J.~Han, L.~Guo, and G.-S. Xia, ``Remote sensing image scene
  classification meets deep learning: Challenges, methods, benchmarks, and
  opportunities,'' \emph{IEEE J. Sel. Topics Appl. Earth Observ. Remote Sens.},
  vol.~13, pp. 3735--3756, 2020.

\bibitem{adv_rs}
Y.~Xu, B.~Du, and L.~Zhang, ``Assessing the threat of adversarial examples on
  deep neural networks for remote sensing scene classification: Attacks and
  defenses,'' \emph{IEEE Trans. Geos. Remote Sens.}, vol.~59, no.~2, pp.
  1604--1617, 2021.

\bibitem{potnis2021semantics}
A.~V. Potnis, S.~S. Durbha, and R.~C. Shinde, ``Semantics-driven remote sensing
  scene understanding framework for grounded spatio-contextual scene
  descriptions,'' \emph{ISPRS Int. J. Geo-Inf.}, vol.~10, no.~1, p.~32, 2021.

\bibitem{sun2021pbnet}
X.~Sun, P.~Wang, C.~Wang, Y.~Liu, and K.~Fu, ``{PBN}et: Part-based
  convolutional neural network for complex composite object detection in remote
  sensing imagery,'' \emph{ISPRS J. Photogramm. Remote Sens.}, vol. 173, pp.
  50--65, 2021.

\bibitem{ghorbanzadeh2022landslide4sense}
O.~Ghorbanzadeh, Y.~Xu, P.~Ghamisi, M.~Kopp, and D.~Kreil, ``Landslide4sense:
  Reference benchmark data and deep learning models for landslide detection,''
  \emph{IEEE Trans. Geos. Remote Sens.}, vol.~60, pp. 1--17, 2022.

\bibitem{ding2020lanet}
L.~Ding, H.~Tang, and L.~Bruzzone, ``{LAN}et: Local attention embedding to
  improve the semantic segmentation of remote sensing images,'' \emph{IEEE
  Trans. Geos. Remote Sens.}, vol.~59, no.~1, pp. 426--435, 2020.

\bibitem{xu2022consistency}
Y.~Xu and P.~Ghamisi, ``Consistency-regularized region-growing network for
  semantic segmentation of urban scenes with point-level annotations,''
  \emph{IEEE Trans. Image Process.}, vol.~31, pp. 5038--5051, 2022.

\bibitem{zhang2016deep}
L.~Zhang, L.~Zhang, and B.~Du, ``Deep learning for remote sensing data: A
  technical tutorial on the state of the art,'' \emph{IEEE Geosci. Remote Sens.
  Mag.}, vol.~4, no.~2, pp. 22--40, 2016.

\bibitem{maggiori2017can}
E.~Maggiori, Y.~Tarabalka, G.~Charpiat, and P.~Alliez, ``Can semantic labeling
  methods generalize to any city? {The} inria aerial image labeling
  benchmark,'' in \emph{Proc. IEEE Int. Geosci. Remote Sens. Symp.}, 2017, pp.
  3226--3229.

\bibitem{lobry2020rsvqa}
S.~Lobry, D.~Marcos, J.~Murray, and D.~Tuia, ``{RSVQA}: Visual question
  answering for remote sensing data,'' \emph{IEEE Trans. Geos. Remote Sens.},
  vol.~58, no.~12, pp. 8555--8566, 2020.

\bibitem{yuan2022remote}
Z.~Yuan, W.~Zhang, C.~Tian, X.~Rong, Z.~Zhang, H.~Wang, K.~Fu, and X.~Sun,
  ``Remote sensing cross-modal text-image retrieval based on global and local
  information,'' \emph{IEEE Trans. Geosci. Remote Sens.}, vol.~60, pp. 1--16,
  2022.

\bibitem{zhang2019description}
X.~Zhang, X.~Wang, X.~Tang, H.~Zhou, and C.~Li, ``Description generation for
  remote sensing images using attribute attention mechanism,'' \emph{Remote
  Sens.}, vol.~11, no.~6, p. 612, 2019.

\bibitem{li2020multi}
Y.~Li, S.~Fang, L.~Jiao, R.~Liu, and R.~Shang, ``A multi-level attention model
  for remote sensing image captions,'' \emph{Remote Sens.}, vol.~12, no.~6, p.
  939, 2020.

\bibitem{liu2022remote}
C.~Liu, R.~Zhao, and Z.~Shi, ``Remote-sensing image captioning based on
  multilayer aggregated transformer,'' \emph{IEEE Geosci. Remote Sens. Lett.},
  vol.~19, pp. 1--5, 2022.

\bibitem{hoxha2022change}
G.~Hoxha, S.~Chouaf, F.~Melgani, and Y.~Smara, ``Change captioning: A new
  paradigm for multitemporal remote sensing image analysis,'' \emph{IEEE Trans.
  Geosci. Remote Sens.}, vol.~60, pp. 1--14, 2022.

\bibitem{liu2022remote_change}
C.~Liu, R.~Zhao, H.~Chen, Z.~Zou, and Z.~Shi, ``Remote sensing image change
  captioning with dual-branch transformers: A new method and a large scale
  dataset,'' \emph{IEEE Trans. Geosci. Remote Sens.}, vol.~60, pp. 1--20, 2022.

\bibitem{chang2023changes}
S.~Chang and P.~Ghamisi, ``Changes to captions: An attentive network for remote
  sensing change captioning,'' \emph{arXiv preprint arXiv:2304.01091}, 2023.

\bibitem{shi2017can}
Z.~Shi and Z.~Zou, ``Can a machine generate humanlike language descriptions for
  a remote sensing image?'' \emph{IEEE Trans. Geos. Remote Sens.}, vol.~55,
  no.~6, pp. 3623--3634, 2017.

\bibitem{lu2017exploring}
X.~Lu, B.~Wang, X.~Zheng, and X.~Li, ``Exploring models and data for remote
  sensing image caption generation,'' \emph{IEEE Trans. Geos. Remote Sens.},
  vol.~56, no.~4, pp. 2183--2195, 2017.

\bibitem{li2020truncation}
X.~Li, X.~Zhang, W.~Huang, and Q.~Wang, ``Truncation cross entropy loss for
  remote sensing image captioning,'' \emph{IEEE Trans. Geos. Remote Sens.},
  vol.~59, no.~6, pp. 5246--5257, 2020.

\bibitem{hoxha2021novel}
G.~Hoxha and F.~Melgani, ``A novel {SVM}-based decoder for remote sensing image
  captioning,'' \emph{IEEE Trans. Geosci. Remote Sens.}, vol.~60, pp. 1--14,
  2021.

\bibitem{zia2022transforming}
U.~Zia, M.~M. Riaz, and A.~Ghafoor, ``Transforming remote sensing images to
  textual descriptions,'' \emph{Int. J. Appl. Earth Obs. Geoinf.}, vol. 108, p.
  102741, 2022.

\bibitem{chen2021remote}
C.~Chen, H.~Ma, G.~Yao, N.~Lv, H.~Yang, C.~Li, and S.~Wan, ``Remote sensing
  image augmentation based on text description for waterside change
  detection,'' \emph{Remote Sens.}, vol.~13, no.~10, p. 1894, 2021.

\bibitem{singh2021sigan}
A.~Singh and L.~Bruzzone, ``{SIGAN}: Spectral index generative adversarial
  network for data augmentation in multispectral remote sensing images,''
  \emph{IEEE Geosci. Remote Sens. Lett.}, vol.~19, pp. 1--5, 2021.

\bibitem{kim2022gan}
J.-H. Kim and Y.~Hwang, ``{GAN}-based synthetic data augmentation for infrared
  small target detection,'' \emph{IEEE Trans. Geosci. Remote Sens.}, vol.~60,
  pp. 1--12, 2022.

\bibitem{bejiga2019retro}
M.~B. Bejiga, F.~Melgani, and A.~Vascotto, ``Retro-remote sensing: Generating
  images from ancient texts,'' \emph{IEEE J. Sel. Topics Appl. Earth Observ.
  Remote Sens.}, vol.~12, no.~3, pp. 950--960, 2019.

\bibitem{zhao2021text}
R.~Zhao and Z.~Shi, ``Text-to-remote-sensing-image generation with structured
  generative adversarial networks,'' \emph{IEEE Geosci. Remote Sens. Lett.},
  vol.~19, pp. 1--5, 2021.

\bibitem{Ramsauer:21}
H.~Ramsauer, B.~Sch\"{a}fl, J.~Lehner, P.~Seidl, M.~Widrich, L.~Gruber,
  M.~Holzleitner, M.~Pavlovi{\'c}, G.~K. Sandve, V.~Greiff, D.~Kreil, M.~Kopp,
  G.~Klambauer, J.~Brandstetter, and S.~Hochreiter, ``{Hopfield} networks is
  all you need,'' in \emph{Proc. Int. Conf. Learn. Representations}, 2021.

\bibitem{reed2016generative}
S.~Reed, Z.~Akata, X.~Yan, L.~Logeswaran, B.~Schiele, and H.~Lee, ``Generative
  adversarial text to image synthesis,'' in \emph{Proc. Int. Conf. Mach.
  Learn.}, 2016, pp. 1060--1069.

\bibitem{reed2016learning}
S.~E. Reed, Z.~Akata, S.~Mohan, S.~Tenka, B.~Schiele, and H.~Lee, ``Learning
  what and where to draw,'' in \emph{Proc. Neural Inf. Process. Syst.}, 2016.

\bibitem{zhang2017stackgan}
H.~Zhang, T.~Xu, H.~Li, S.~Zhang, X.~Wang, X.~Huang, and D.~N. Metaxas,
  ``Stackgan: Text to photo-realistic image synthesis with stacked generative
  adversarial networks,'' in \emph{Proc. IEEE Int. Conf. Comput. Vis.}, 2017,
  pp. 5907--5915.

\bibitem{brown2020language}
T.~Brown, B.~Mann, N.~Ryder, M.~Subbiah, J.~D. Kaplan, P.~Dhariwal,
  A.~Neelakantan, P.~Shyam, G.~Sastry, A.~Askell \emph{et~al.}, ``Language
  models are few-shot learners,'' in \emph{Proc. Neural Inf. Process. Syst.},
  vol.~33, 2020, pp. 1877--1901.

\bibitem{ramesh2021zero}
A.~Ramesh, M.~Pavlov, G.~Goh, S.~Gray, C.~Voss, A.~Radford, M.~Chen, and
  I.~Sutskever, ``Zero-shot text-to-image generation,'' in \emph{Proc. Int.
  Conf. Mach. Learn.}, 2021, pp. 8821--8831.

\bibitem{ramesh2022hierarchical}
A.~Ramesh, P.~Dhariwal, A.~Nichol, C.~Chu, and M.~Chen, ``Hierarchical
  text-conditional image generation with clip latents,'' \emph{arXiv preprint
  arXiv:2204.06125}, 2022.

\bibitem{radford2021learning}
A.~Radford, J.~W. Kim, C.~Hallacy, A.~Ramesh, G.~Goh, S.~Agarwal, G.~Sastry,
  A.~Askell, P.~Mishkin, J.~Clark \emph{et~al.}, ``Learning transferable visual
  models from natural language supervision,'' in \emph{Proc. Int. Conf. Mach.
  Learn.}, 2021, pp. 8748--8763.

\bibitem{liu2001evaluation}
C.-L. Liu and M.~Nakagawa, ``Evaluation of prototype learning algorithms for
  nearest-neighbor classifier in application to handwritten character
  recognition,'' \emph{Pattern Recognit.}, vol.~34, no.~3, pp. 601--615, 2001.

\bibitem{dudani1976distance}
S.~A. Dudani, ``The distance-weighted k-nearest-neighbor rule,'' \emph{IEEE
  Trans. Syst. Man Cybern.}, no.~4, pp. 325--327, 1976.

\bibitem{kohonen1990self}
T.~Kohonen, ``The self-organizing map,'' \emph{Proc. IEEE}, vol.~78, no.~9, pp.
  1464--1480, 1990.

\bibitem{li2021adaptive}
G.~Li, V.~Jampani, L.~Sevilla-Lara, D.~Sun, J.~Kim, and J.~Kim, ``Adaptive
  prototype learning and allocation for few-shot segmentation,'' in \emph{Proc.
  IEEE Conf. Comput. Vis. Pattern Recognit.}, 2021, pp. 8334--8343.

\bibitem{yang2018robust}
H.-M. Yang, X.-Y. Zhang, F.~Yin, and C.-L. Liu, ``Robust classification with
  convolutional prototype learning,'' in \emph{Proc. IEEE Conf. Comput. Vis.
  Pattern Recognit.}, 2018, pp. 3474--3482.

\bibitem{dong2018few}
N.~Dong and E.~P. Xing, ``Few-shot semantic segmentation with prototype
  learning.'' in \emph{BMVC}, vol.~3, no.~4, 2018.

\bibitem{wang2021interactive}
X.~Wang, L.~Zhu, H.~Wang, and Y.~Yang, ``Interactive prototype learning for
  egocentric action recognition,'' in \emph{Proc. IEEE Conf. Comput. Vis.
  Pattern Recognit.}, 2021, pp. 8168--8177.

\bibitem{deng2021variational}
J.~Deng, J.~Guo, J.~Yang, A.~Lattas, and S.~Zafeiriou, ``Variational prototype
  learning for deep face recognition,'' in \emph{Proc. IEEE Conf. Comput. Vis.
  Pattern Recognit.}, 2021, pp. 11\,906--11\,915.

\bibitem{mceliece1987capacity}
R.~McEliece, E.~Posner, E.~Rodemich, and S.~Venkatesh, ``The capacity of the
  {Hopfield} associative memory,'' \emph{IEEE Trans. Inf. Theory}, vol.~33,
  no.~4, pp. 461--482, 1987.

\bibitem{Hopfield:82}
J.~J. Hopfield, ``Neural networks and physical systems with emergent collective
  computational abilities,'' \emph{Proc. Nat. Acad. Sci.}, vol.~79, no.~8, pp.
  2554--2558, 1982.

\bibitem{Hopfield:84}
------, ``Neurons with graded response have collective computational properties
  like those of two-state neurons,'' \emph{Proc. Nat. Acad. Sci.}, vol.~81,
  no.~10, pp. 3088--3092, 1984.

\bibitem{Krotov:16}
D.~Krotov and J.~J. Hopfield, ``Dense associative memory for pattern
  recognition,'' in \emph{Proc. Neural Inf. Process. Syst.}, D.~D. Lee,
  M.~Sugiyama, U.~V. Luxburg, I.~Guyon, and R.~Garnett, Eds.\hskip 1em plus
  0.5em minus 0.4em\relax Curran Associates, Inc., 2016, pp. 1172--1180.

\bibitem{Demircigil:17}
M.~Demircigil, J.~Heusel, M.~L{\"{o}}we, S.~Upgang, and F.~Vermet, ``On a model
  of associative memory with huge storage capacity,'' \emph{Journal of
  Statistical Physics}, vol. 168, no.~2, pp. 288--299, 2017.

\bibitem{widrich2020modern}
M.~Widrich, B.~Sch{\"a}fl, M.~Pavlovi{\'c}, H.~Ramsauer, L.~Gruber,
  M.~Holzleitner, J.~Brandstetter, G.~K. Sandve, V.~Greiff, S.~Hochreiter
  \emph{et~al.}, ``Modern {Hopfield} networks and attention for immune
  repertoire classification,'' in \emph{Proc. Neural Inf. Process. Syst.},
  vol.~33, 2020, pp. 18\,832--18\,845.

\bibitem{Seidl:22}
P.~Seidl, P.~Renz, N.~Dyubankova, P.~Neves, J.~Verhoeven, J.~K. Wegner,
  M.~Segler, S.~Hochreiter, and G.~Klambauer, ``Improving few-and zero-shot
  reaction template prediction using modern {Hopfield} networks,'' \emph{J.
  Chem. Inf. Model.}, 2022.

\bibitem{furst2021cloob}
A.~F{\"u}rst, E.~Rumetshofer, J.~Lehner, V.~T. Tran, F.~Tang, H.~Ramsauer,
  D.~Kreil, M.~Kopp, G.~Klambauer, A.~Bitto \emph{et~al.}, ``{CLOOB: Modern
  Hopfield networks with InfoLOOB outperform CLIP},'' in \emph{Proc. Neural
  Inf. Process. Syst.}, vol.~35, 2022, pp. 20\,450--20\,468.

\bibitem{sennrich2015neural}
R.~Sennrich, B.~Haddow, and A.~Birch, ``Neural machine translation of rare
  words with subword units,'' in \emph{Proc. Annu. Meeting Assoc. Comput.
  Linguistics}, 2016, pp. 1715--1725.

\bibitem{vaswani2017attention}
A.~Vaswani, N.~Shazeer, N.~Parmar, J.~Uszkoreit, L.~Jones, A.~N. Gomez,
  {\L}.~Kaiser, and I.~Polosukhin, ``Attention is all you need,'' in
  \emph{Proc. Neural Inf. Process. Syst.}, vol.~30, 2017.

\bibitem{gehring2017convolutional}
J.~Gehring, M.~Auli, D.~Grangier, D.~Yarats, and Y.~N. Dauphin, ``Convolutional
  sequence to sequence learning,'' in \emph{Proc. Int. Conf. Mach. Learn.},
  2017, pp. 1243--1252.

\bibitem{graves2013generating}
A.~Graves, ``Generating sequences with recurrent neural networks,'' \emph{arXiv
  preprint arXiv:1308.0850}, 2013.

\bibitem{van2017neural}
A.~Van Den~Oord, O.~Vinyals \emph{et~al.}, ``Neural discrete representation
  learning,'' in \emph{Proc. Neural Inf. Process. Syst.}, 2017.

\bibitem{esser2021taming}
P.~Esser, R.~Rombach, and B.~Ommer, ``Taming transformers for high-resolution
  image synthesis,'' in \emph{Proc. IEEE Conf. Comput. Vis. Pattern Recognit.},
  2021, pp. 12\,873--12\,883.

\bibitem{shaw2018self}
P.~Shaw, J.~Uszkoreit, and A.~Vaswani, ``Self-attention with relative position
  representations,'' \emph{arXiv preprint arXiv:1803.02155}, 2018.

\bibitem{xu2021self}
Y.~Xu, B.~Du, and L.~Zhang, ``Self-attention context network: Addressing the
  threat of adversarial attacks for hyperspectral image classification,''
  \emph{IEEE Trans. Image Process.}, vol.~30, pp. 8671--8685, 2021.

\bibitem{salimans2016improved}
T.~Salimans, I.~Goodfellow, W.~Zaremba, V.~Cheung, A.~Radford, and X.~Chen,
  ``Improved techniques for training {GANs},'' in \emph{Proc. Neural Inf.
  Process. Syst.}, vol.~29, 2016.

\bibitem{heusel2017gans}
M.~Heusel, H.~Ramsauer, T.~Unterthiner, B.~Nessler, and S.~Hochreiter, ``Gans
  trained by a two time-scale update rule converge to a local nash
  equilibrium,'' in \emph{Proc. Neural Inf. Process. Syst.}, vol.~30, 2017.

\bibitem{inception}
C.~Szegedy, V.~Vanhoucke, S.~Ioffe, J.~Shlens, and Z.~Wojna, ``Rethinking the
  inception architecture for computer vision,'' in \emph{Proc. IEEE Conf.
  Comput. Vis. Pattern Recognit.}, 2016, pp. 2818--2826.

\bibitem{saharia2022photorealistic}
C.~Saharia, W.~Chan, S.~Saxena, L.~Li, J.~Whang, E.~L. Denton, K.~Ghasemipour,
  R.~Gontijo~Lopes, B.~Karagol~Ayan, T.~Salimans \emph{et~al.},
  ``Photorealistic text-to-image diffusion models with deep language
  understanding,'' in \emph{Proc. Neural Inf. Process. Syst.}, 2022.

\bibitem{resnet}
K.~He, X.~Zhang, S.~Ren, and J.~Sun, ``Deep residual learning for image
  recognition,'' in \emph{Proc. IEEE Conf. Comput. Vis. Pattern Recognit.},
  2016, pp. 770--778.

\bibitem{xu2018attngan}
T.~Xu, P.~Zhang, Q.~Huang, H.~Zhang, Z.~Gan, X.~Huang, and X.~He, ``Attngan:
  Fine-grained text to image generation with attentional generative adversarial
  networks,'' in \emph{Proc. IEEE Conf. Comput. Vis. Pattern Recognit.}, 2018,
  pp. 1316--1324.

\bibitem{ruan2021dae}
S.~Ruan, Y.~Zhang, K.~Zhang, Y.~Fan, F.~Tang, Q.~Liu, and E.~Chen, ``{DAE-GAN}:
  Dynamic aspect-aware gan for text-to-image synthesis,'' in \emph{Proc. IEEE
  Int. Conf. Comput. Vis.}, 2021, pp. 13\,960--13\,969.

\bibitem{tao2022df}
M.~Tao, H.~Tang, F.~Wu, X.-Y. Jing, B.-K. Bao, and C.~Xu, ``{DF-GAN}: A simple
  and effective baseline for text-to-image synthesis,'' in \emph{Proc. IEEE
  Conf. Comput. Vis. Pattern Recognit.}, 2022, pp. 16\,515--16\,525.

\bibitem{zhou2022towards}
Y.~Zhou, R.~Zhang, C.~Chen, C.~Li, C.~Tensmeyer, T.~Yu, J.~Gu, J.~Xu, and
  T.~Sun, ``Towards language-free training for text-to-image generation,'' in
  \emph{Proc. IEEE Conf. Comput. Vis. Pattern Recognit.}, 2022, pp.
  17\,907--17\,917.

\end{thebibliography}

\end{document}